\documentclass[letterpaper]{article}
\usepackage{aaai2026}
\usepackage{times}  % DO NOT CHANGE THIS
\usepackage{helvet}  % DO NOT CHANGE THIS
\usepackage{courier}  % DO NOT CHANGE THIS
\usepackage[hyphens]{url}  % DO NOT CHANGE THIS
\usepackage{graphicx} % DO NOT CHANGE THIS
\usepackage{natbib}  % DO NOT CHANGE THIS AND DO NOT ADD ANY OPTIONS TO IT
\usepackage{caption} % DO NOT CHANGE THIS AND DO NOT ADD ANY OPTIONS TO IT
\usepackage{algorithm}
\usepackage{algorithmic}
\usepackage{amsmath}
\usepackage{amssymb}
\usepackage{pifont}
\usepackage{tikz}
\usepackage{multirow}
\usepackage{booktabs}
\usepackage{makecell}
\usepackage{microtype}
\usepackage{arydshln}
\usepackage{relsize} 
\usepackage[table]{xcolor}
\usepackage{array}
\usepackage{amsmath,bm}

\usepackage{newfloat}
\usepackage{listings}
\DeclareCaptionStyle{ruled}{labelfont=normalfont,labelsep=colon,strut=off} % DO NOT CHANGE THIS
\floatstyle{ruled}
\newfloat{listing}{tb}{lst}{}
\floatname{listing}{Listing}
\title{PhysioME: A Robust Multimodal Self-Supervised Framework \\ for Physiological Signals with Missing Modalities}
\author{
    Cheol-Hui Lee\textsuperscript{\rm 1,2\textdagger}, Hwa-Yeon Lee\textsuperscript{\rm 1,2\textdagger}, Min-Kyung Jung\textsuperscript{\rm 1,2}, Dong-Joo Kim\textsuperscript{\rm 1,2,3}\textsuperscript{*}
}
\affiliations{
    \textsuperscript{\rm 1}Department of Brain and Cognitive Engineering, Korea University, Seoul, South Korea\\
    \textsuperscript{\rm 2}Interdisciplinary Program in Precision Public Health, Korea University, Seoul, South Korea\\
    \textsuperscript{\rm 3}Department of Neurology, Korea University College of Medicine, Seoul, South Korea\\
    dlcjfgmlnasa28@korea.ac.kr, belita0152@korea.ac.kr, mkjung@korea.ac.kr, dongjookim@korea.ac.kr\\
}

\usepackage{bibentry}
\begin{document}

\maketitle

\begin{abstract}
Missing or corrupted modalities are common in physiological signal-based medical applications owing to hardware constraints or motion artifacts. However, most existing methods assume the availability of all modalities, resulting in substantial performance degradation in the absence of any modality. To overcome this limitation, this study proposes PhysioME, a robust framework designed to ensure reliable performance under missing modality conditions. PhysioME adopts: (1) a multimodal self-supervised learning approach that combines contrastive learning with masked prediction; (2) a Dual-Path-NeuroNet backbone tailored to capture the temporal dynamics of each physiological signal modality; and (3) a restoration decoder that reconstructs missing modality tokens, enabling flexible processing of incomplete inputs. The experimental results show that PhysioME achieves high consistency and generalization performance across various missing modality scenarios. These findings highlight the potential of PhysioME as a reliable tool for supporting clinical decision-making in real-world settings with imperfect data availability.
\end{abstract}

\section{Introduction}
In real-world clinical environments, it is often challenging to continuously acquire all physiological signals because of sensor malfunctions, patient movements, or hardware limitations \cite{iranfar2021relearn}. However, most existing methods are designed under the assumption that all modalities are available, which leads to a performance degradation when some modalities are missing \cite{reza2024robust}. 

To overcome these limitations and enable continuous patient monitoring, specific model architectures that can robustly handle missing modality scenarios are required. There are two main strategies: “dedicated model” and “single-model” approaches \cite{lee2023robustssf, wu2024deep}. The former trains independent networks for each possible combination of modalities. This strategy offers precise handling of specific cases, but suffers from excessive computational and resource demands, thereby limiting its applicability in practice \cite{bachmann2022multimae}. By contrast, the latter handles all scenarios within a unified architecture, thereby greatly improving its practicality in clinical environments. As a result of this advantage, recent studies have increasingly focused on the development of single-model approaches that are capable of handling diverse missing modality scenarios within a unified framework \cite{wu2024deep, bachmann2022multimae, lee2023robustssf, deldari2024crossl, zhao2024maskmentor}.

However, most single-models focus on vision tasks \cite{bachmann2022multimae, lee2023robustssf, zhao2024maskmentor}, restricting the extensibility of using physiological signals or heterogeneous sensors to address structural challenges. First, vision models are designed to process spatial information within images and struggle to capture the temporal continuity of signals \cite{wang2020deep}. Second, simply replacing missing signals with mask or class tokens often fails to capture the semantics of actual missing scenarios, thereby hampering the generalization performance \cite{bachmann2022multimae, zhao2024maskmentor}. Third, most missing modality models are built on supervised learning frameworks that require extensive labeled datasets, which are scarce in the medical domain \cite{esteva2019guide, lee2024neuronet}.

To overcome these limitations, we propose \textbf{PhysioME}, a single-model-based missing modality framework specifically designed for physiological signals. The aim of this study is to achieve robust performance in clinical settings where certain modalities may be missing. PhysioME has the following key characteristics:

\begin{itemize}
    \item PhysioME is a multimodal self-supervised learning (SSL) framework tailored for heterogeneous physiological signals. It integrates contrastive learning and masked prediction to learn discriminative and generative representations jointly, thereby enabling robust and generalizable feature learning for physiological data.
    \item The SSL backbone of PhysioME, \textbf{Dual-Path (DP)-NeuroNet}, effectively encodes the temporal characteristics of physiological signals. It applies distinct temporal augmentations to two parallel paths that share the same NeuroNet \cite{lee2024neuronet} structure, thereby facilitating rich representation learning. The resulting modality-specific representations serve as inputs for PhysioME.
    \item PhysioME incorporates a dedicated \textbf{restoration decoder} module for each modality. These decoders are designed to reconstruct missing modality tokens by leveraging the observed modalities, allowing the model to robustly handle a wide range of missing modality scenarios in physiological signals.
\end{itemize}

\begin{figure*}[!t]
\centering
\includegraphics[width=0.9\textwidth]{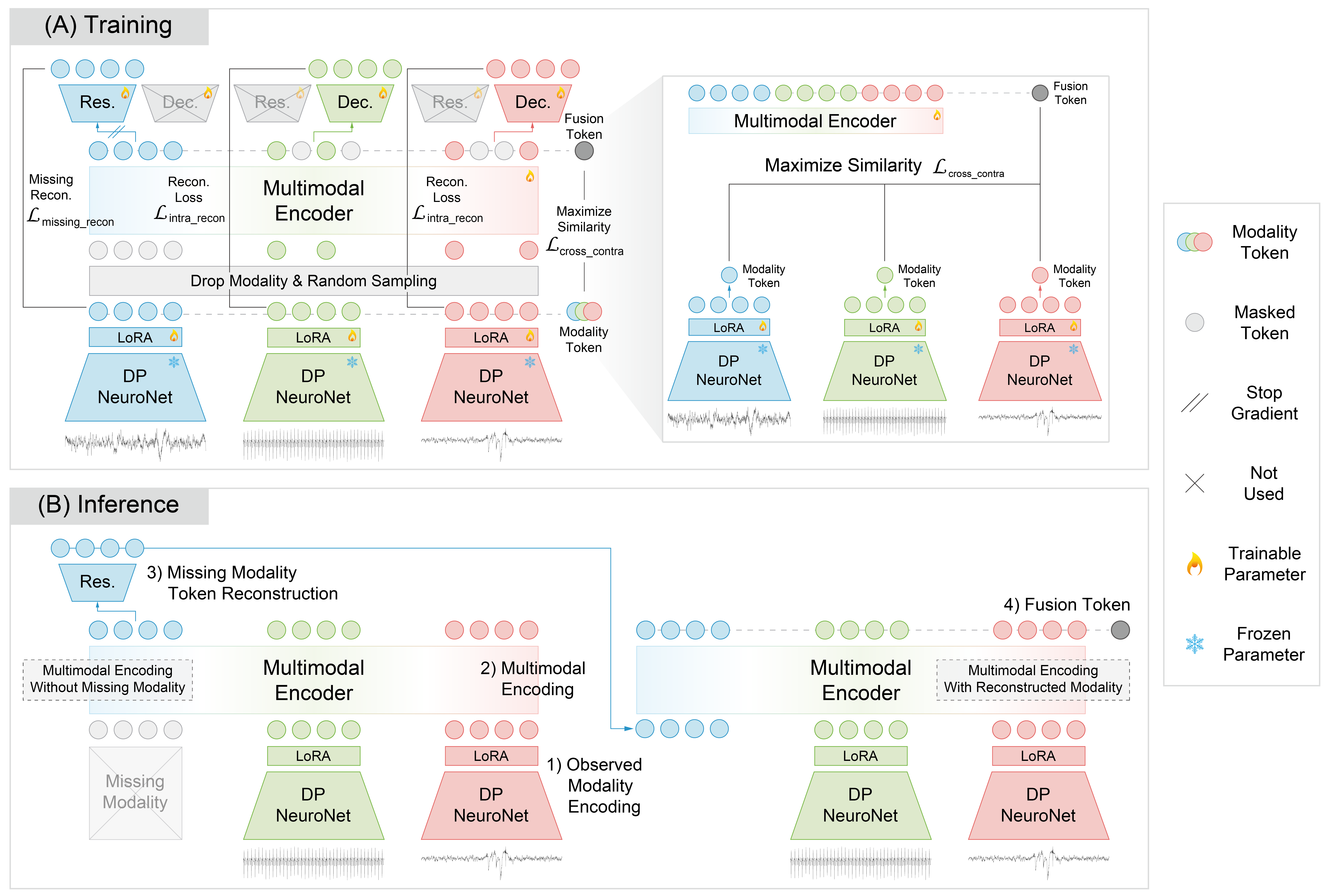}
\caption{Overview of PhysioME architecture. (A) Training workflow of PhysioME: a multimodal SSL framework for missing modalities. (B) Inference procedure using the pretrained PhysioME.}
\label{fig:figure1}
\end{figure*}

This proposed framework consistently maintains high performance under various missing modality conditions. This model demonstrates strong robustness and generalization in sleep stage classification and hypotension prediction using clinical datasets.

\section{Methods: PhysioME}
\subsection{DP-NeuroNet}

\begin{figure}[!t]
    \centerline{\includegraphics[width=1\columnwidth]{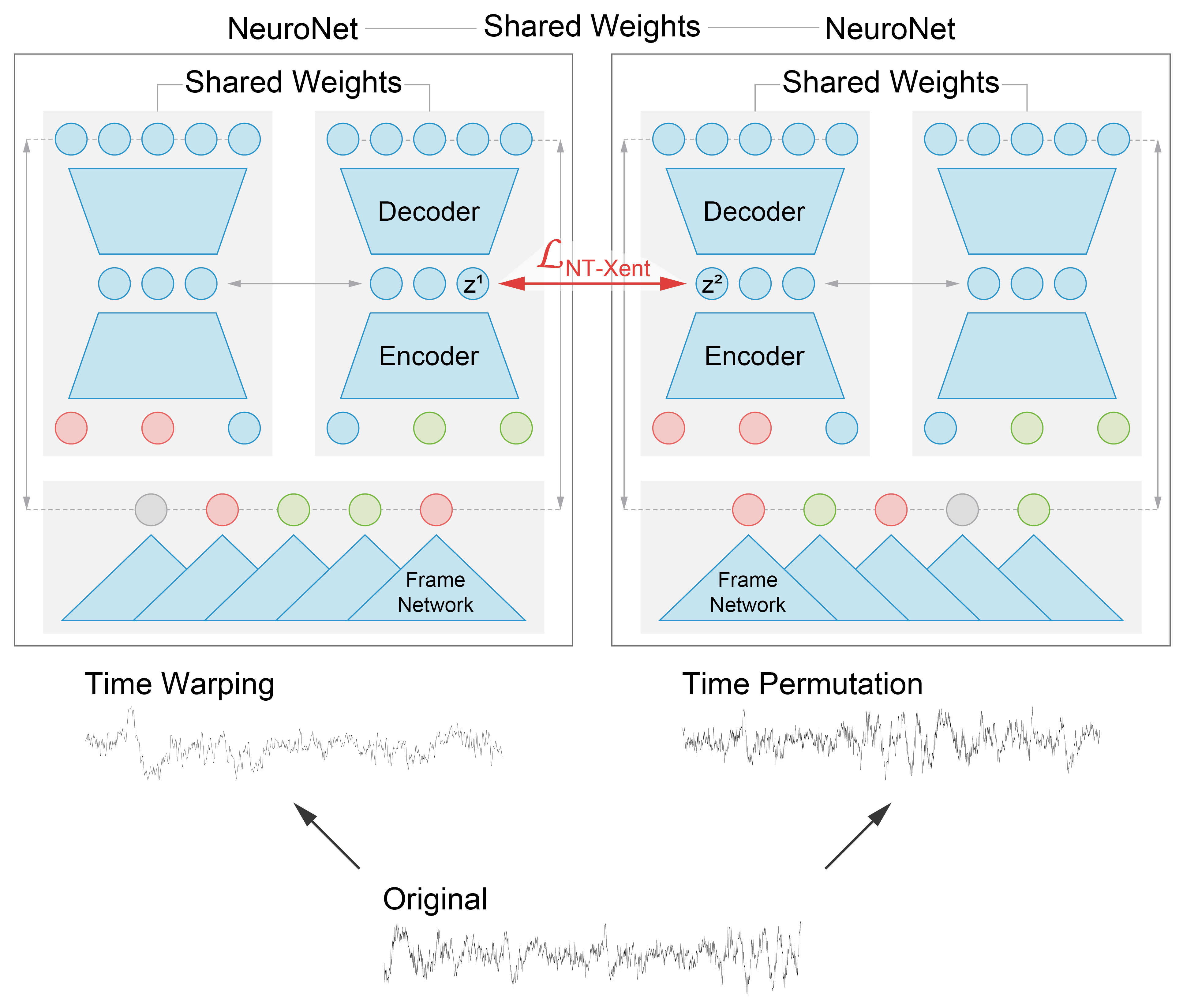}}
    \caption{DP-NeuroNet structure.}
    \label{fig:figure2}
\end{figure}

DP-NeuroNet comprises two identical instances of NeuroNet that share weights (Figure \ref{fig:figure2}). NeuroNet \cite{lee2024neuronet} is an SSL framework for physiological signals that is designed to effectively learn generalizable representations from unlabeled data by combining masked prediction and contrastive learning.

NeuroNet consists of a multi-scale 1D ResNet-based frame network and a vision transformer (ViT)-based encoder-decoder architecture. A subset of the token sequence produced by the frame network is randomly sampled and reconstructed through the encoder-decoder, with the reconstruction optimized using the mean squared error (MSE) loss. In addition, contrastive learning is performed using the NT-Xent loss \cite{sohn2016improved} applied to two independently sampled token sequences. (Appendix I for architectural details of NeuroNet).

Each augmented input, $\tilde{x}_i^{(1)}$ and $\tilde{x}_i^{(2)}$, is independently processed using NeuroNet. The representations $z^{(1)}$ and $z^{(2)}$ are then extracted using the NeuroNet encoder $f_{\text{neuronet\_enc}}$ (composed of the frame network and ViT encoder), followed by a multi-layer perceptron (MLP) layer. NT-Xent loss \cite{sohn2016improved} is then applied to these representations to encourage similarities between different augmented views of the same input. The formula is as follows: 
\begin{equation}
\begin{aligned}
\mathcal{L}_{\mathrm{NT\text{-}Xent}} &= \frac{1}{2B} \sum_{b=1}^{B} \left[
    \ell\left(z_b^{(1)}, z_b^{(2)}\right) + \ell\left(z_b^{(2)}, z_b^{(1)}\right)
\right] 
\end{aligned}
\end{equation}
\begin{equation}
\begin{aligned}
\ell(z_a, z_b) &= -\log \frac{
    \exp\left(\mathrm{sim}(z_a, z_b)/\tau\right)
}{
    \sum_{k=1}^{2B} \mathbf{1}_{[k \neq a]} \exp\left(\mathrm{sim}(z_a, z_k)/\tau\right)
}
\end{aligned}
\end{equation} where $B$ denotes the batch size and $z_1^{(1)}$ refers to the first sample in the batch from $z^{(1)}$. The function $\text{sim}(\cdot)$ represents the cosine similarity.

\subsection{PhysioME} PhysioME is a multimodal SSL framework that is specifically designed to handle missing modalities (Figure \ref{fig:figure1}). Its core components are as follows:

\paragraph{(A)\ Training}

\subsubsection{Modality Encoder} Each physiological signal is encoded using the pre-trained encoder $f_{\text{neuronet\_enc}}$  from DP-NeuroNet, which serves as the modality encoder. A total of $M$ modality encoders are used, each responsible for extracting features specific to the $m$-th modality, where $m \in \{1, 2, \ldots, M\}$. Given an input signal $x^{(m)}$, the corresponding modality encoder produces a token sequence $e^{(m)} = \{ e_{i}^{(m)} \}_{i=1}^{N} = f_{\text{modality\_enc}}^{(m)}(x^{(m)}) \equiv f_{\text{neuronet\_enc}}^{(m)}(x^{(m)})$, where $N$ denotes the length of the token sequence. To adapt the modality encoders for PhysioME, low-rank adaptation (LoRA) \cite{hu2022lora} is applied to a subset of layers, whereas the remaining layers are frozen during training.

\subsubsection{Multimodal Encoder} Each $M$ modality encoder produces a token sequence $e^{(m)}$, which is then fed into a ViT-based multimodal encoder to generate a unified sequence of tokens. This process consists of three main stages.

\begin{enumerate}
    \item \textbf{Positional and Modality Encoding}: Each token sequence $e^{(m)}$ is passed through an MLP layer, after which positional encoding $pe$ and modality encoding $mt^{(m)}$ are added. This results in a new token sequence: $z^{(m)} = \text{MLP}(e^{(m)}) + pe + mt^{(m)}$.

    \item \textbf{Drop Modality and Random Sampling}: For each of the $M$ token sequences $z^{(m)}$, either modality dropping or random sampling is randomly selected and applied.

        \begin{itemize}
            \item \textbf{Drop modality} simulates a missing modality scenario by removing the entire token sequence $z^{(m)}$ and replacing it with a mask token $\mu$.
 
            \item \textbf{Random sampling} involves the selection of $\tilde{N}$ tokens from $z^{(m)}$, where $\tilde{N} < N$. The selected subset token sequence ${\tilde{z}}^{(m)}$ is constructed based on an index set $\mathcal{H} \subseteq \{1, \ldots, N\}$.

        \end{itemize}
    \item \textbf{Fusion Token}: The token sequence $\tilde{h}$ is input into the multimodal encoder $f_{\text{multimodal\_enc}}$ to generate the sampled fusion token sequence $\tilde{o}$.

\end{enumerate}

\subsubsection{Modality Decoder} The ViT-based modality decoders are constructed independently for each modality. Each decoder takes only a subset of tokens $\tilde{o}^{(m)}$ corresponding to the set of modalities $\tilde{\mathcal{M}}$ that are not subjected to drop modality, extracted from the output $\tilde{o}$ of the multimodal encoder.

In this process, mask tokens $\mu$ are inserted into ${\tilde{o}}^{(m)}$ at the positions corresponding to the token indices not included in the random sampling subset $\mathcal{H}$. The resulting sequence is then passed through the modality decoder $f_{\text{modality\_dec}}^{(m)}$ and an MLP layer to generate a new token sequence: $d^{(m)} = \text{MLP}(f_{\text{modality\_dec}}^{(m)}(\mu,\, {\tilde{o}}^{(m)} \mid m \in \tilde{\mathcal{M}}))$. The output $d^{(m)}$ is used to reconstruct the original modality encoder output $e^{(m)}$ , and this reconstruction is optimized using the MSE loss. After SSL is completed, the modality decoders are discarded.

\subsubsection{Restoration Decoder} The restoration decoder adopts a ViT-based architecture similar to the modality decoder but is designed with a deeper structure to enable more precise reconstruction. During training, a stop-gradient operation is applied to block backpropagation to the multimodal encoder and other networks, allowing the restoration decoder to be trained independently with the sole focus on reconstruction.

The restoration decoder  $f_{\text{restoration\_dec}}^{(m)}$ takes only the modality-specific tokens $\tilde{o}^{(m)}$ corresponding to the modalities where the drop modality is applied, extracted from the output $\tilde{o}$ of the multimodal encoder. It then generates a new token sequence: $g^{(m)} = \text{MLP}(f_{\text{restoration\_dec}}^{(m)}({\tilde{o}}^{(m)} \mid m \in \mathcal{M} \setminus \tilde{\mathcal{M}}))$, which is trained to reconstruct the original modality encoder output $e^{(m)}$ using the MSE loss. Once the training is complete, the restoration decoder is utilized during inference to estimate the token sequence corresponding to the missing modalities.

\subsubsection{Objective Loss}

\begin{enumerate}
    \item \textbf{Reconstruction Loss}: The output $d^{(m)}$ from the modality decoder is defined only for modalities $m \in \tilde{\mathcal{M}}$ where drop modality is not applied. To reconstruct the corresponding modality encoder output $e^{(m)}$, the MSE loss is employed. Reconstruction is performed only on the token indices that were not selected during random sampling, denoted by the index set ${\bar{\mathcal{H}}}^{(m)}$. The loss is formulated as follows: 
    
    \begin{equation}
    \begin{aligned}
        \mathcal{L}_{\text{intra\_recon}} = \frac{1}{|\tilde{\mathcal{M}}|} \sum_{m \in \tilde{\mathcal{M}}} \frac{1}{|\overline{\mathcal{H}}(m)|} \sum_{i \in \overline{\mathcal{H}}(m)} \left\| d_i^{(m)} - e_i^{(m)} \right\|_2^2
    \end{aligned}
    \end{equation}

    \item \textbf{Missing Modality Reconstruction Loss}: The restoration decoder is trained only for the dropped modalities, i.e., $m \in \mathcal{M} \setminus \tilde{\mathcal{M}}$. The output $g^{(m)}$ is optimized to reconstruct the corresponding modality encoder output $e^{(m)}$. The loss is formulated as follows: \begin{equation}
        \mathcal{L}_{\text{missing\_recon}} = \frac{1}{|\mathcal{M} \setminus \tilde{\mathcal{M}}|} \sum_{m \in \mathcal{M} \setminus \tilde{\mathcal{M}}} \frac{1}{N} \sum_{i=1}^{N} \left\| g_i^{(m)} - e_i^{(m)} \right\|_2^2
    \end{equation}

    \item \textbf{Cross-Modality Contrastive Loss}: To align the representations between the modality and multimodal encoders, contrastive learning based on the NT-Xent loss is applied. This encourages the model to learn the relationship between the single-modal representation $e^{(m)}$, extracted from each modality encoder, and the fused representation $o$ obtained from the multimodal encoder. Consequently, semantic consistency across modalities is preserved, and the quality of the fused representation is enhanced.

    \hspace*{1em} Specifically, the representations $e^{(m)}$ and $o$ are averaged along the sequence dimension, normalized, and passed through an MLP layer to obtain ${em}^{(m)}$ and $om$, respectively. The NT-Xent loss is computed based on the similarity between these two representations.
    \begin{equation}
    \begin{aligned}
        \mathcal{L}_{\text{NT-Xent}}^{(m)} = \frac{1}{2B} \sum_{b=1}^{B} \left[ l\left(\textit{em}_b^{(m)}, \textit{om}_b\right) + l\left(\textit{om}_b, \textit{em}_b^{(m)}\right) \right]
    \end{aligned}
    \end{equation} \begin{equation}
    \begin{aligned}
    l(z_a, z_b) = -\log \frac{\exp\left(\text{sim}(z_a, z_b)/\tau\right)}{\sum_{k=1}^{2B} \mathbf{1}_{[k \neq a]} \exp\left(\text{sim}(z_a, z_k)/\tau\right)}
    \end{aligned}
    \end{equation} where $B$ denotes the batch size, while $\text{sim}(\cdot)$ and $\tau$ represent the cosine similarity and temperature parameter, respectively. The final cross-modality contrastive loss is defined as the average loss across all modalities. The formula is as follows: \begin{equation}
    \begin{aligned}
    \mathcal{L}_{\text{cross\_contra}} = \frac{1}{|\mathcal{M}|} \sum_{m \in \mathcal{M}} \mathcal{L}_{\text{NT-Xent}}^{(m)}
    \end{aligned}
    \end{equation}

    \item \textbf{Final Loss}: The model is optimized using a combined objective that integrates the three loss components as follows: \begin{equation}
    \mathcal{L} = \alpha \mathcal{L}_{\mathrm{intra\_recon}} + \beta \mathcal{L}_{\mathrm{missing\_recon}} + \gamma \mathcal{L}_{\mathrm{cross\_contra}}
    \end{equation} where $\alpha$, $\beta$, and $\gamma$ are hyperparameters that balance the contributions of the three losses.
\end{enumerate}

\paragraph{(B)\ Inference} To handle scenarios in which certain modalities are missing, PhysioME follows the inference procedure as follows:

\subsubsection{Observed Modality Encoding} For each observed modality $\mathcal{M}_{\text{obs}} \in \{1, \ldots, M\}$, the input $x^{(m)}$ is passed through the modality encoder $f_{\text{modality\_enc}}^{(m)}$  and an MLP layer, followed by the addition of positional encoding $pe$ and a modality token $mt^{(m)}$.
\begin{equation} 
    e^{(m)} = f_{\text{modality\_enc}}^{(m)}(x^{(m)}), \quad \text{for } m \in \mathcal{M}_{\text{obs}} \\
\end{equation} 
\begin{equation} 
    z^{(m)} = \{ z_i^{(m)} \}_{i=1}^{N} = \text{MLP}(e^{(m)}) + pe + mt^{(m)}
\end{equation} 

\subsubsection{Multimodal Encoding} The token sequences $z^{(m)}$ are merged with the masked tokens $\mu$ and fed into the multimodal encoder $f_{\text{multimodal\_enc}}$. \begin{equation} 
\begin{aligned}
    o = f_{\text{multimodal\_enc}}(\text{concat}(\mu,\ z^{(m)} \mid m \in \mathcal{M}_{\text{obs}}))
\end{aligned}
\end{equation}

\subsubsection{Missing Modality Token Reconstruction} From the output $o$ of $f_{\text{multimodal\_enc}}$, the token sequences $o^{(m)}$ corresponding to the unobserved modalities $\mathcal{M}_{\text{miss}} \subset \{1, \ldots, M\} \setminus \mathcal{M}_{\text{obs}}$ are extracted. These are then passed through the restoration decoder $f_{\text{restoration\_dec}}^{(m)}$ and an MLP layer to reconstruct the sequence tokens $g^{(m)}$ for the missing modalities.
\begin{equation} 
g^{(m)} = \text{MLP}(f_{\text{restoration\_dec}}^{(m)}(o^{(m)})), \quad \text{for } m \in \mathcal{M}_{\text{miss}}
\end{equation} 

\subsubsection{Generating a Fusion Token} 
The token sequences reconstructed by the restoration decoder are merged with those extracted from the modality encoders, and then fed into the multimodal encoder. The resulting output token sequence $ft$ includes a class token that represents the entire sequence and serves as a global representation for the final prediction in the downstream tasks. \begin{equation}
\begin{aligned}
ft = f_{\text{multimodal\_enc}}(\text{concat}(&g^{(m)} \mid m \in \mathcal{M}_{\text{miss}}, \\
                                              &e^{(m)} \mid m \in \mathcal{M}_{\text{obs}}))
\end{aligned}
\end{equation}

%%%%%%%%%%%%%%%%%%%%%%%%%%%%%%%%%%%%%%%%%%%%%%%%%%%%%%%%%%%%%%%%%%%%%%%%%%%%%%%%%%%%%%%%%%%%%%%%%%%%%%%%%%%%%%%%%%%%%%%%%%%%%%%%%%%%%%%%%
% Table 1
\begin{table*}[!t]
\centering
\footnotesize
\def\arraystretch{0.8}
\setlength{\tabcolsep}{3.6pt}
\begin{tabular}{
>{\centering\arraybackslash}m{1.2cm} 
>{\centering\arraybackslash}m{1.2cm} 
>{\centering\arraybackslash}m{1.0cm} 
cc cc cc cc cc cc cc}
\toprule
\multicolumn{3}{c}{\textit{Modality}} & \multicolumn{4}{c}{\textit{Dedicated Model}} & \multicolumn{10}{c}{\textit{Single Model}} \\
\cmidrule(lr){1-3} \cmidrule(lr){4-7} \cmidrule(lr){8-17}
& & 
& \multicolumn{2}{c}{ContraWR} & \multicolumn{2}{c}{SynthSleepNet} 
& \multicolumn{2}{c}{MultiMAE} & \multicolumn{2}{c}{RobustSsF} 
& \multicolumn{2}{c}{CroSSL} & \multicolumn{2}{c}{MaskMentor} 
& \multicolumn{2}{c}{PhysioME} \\
\cmidrule(lr){4-5} \cmidrule(lr){6-7} \cmidrule(lr){8-9}
\cmidrule(lr){10-11} \cmidrule(lr){12-13} \cmidrule(lr){14-15} \cmidrule(lr){16-17}
\multirow{-3}{*}{\shortstack{\\ {EEG} \\{Fpz-Cz}}} 
& \multirow{-3}{*}{\shortstack{\\ {EEG} \\{Pz-Cz}}} 
& \multirow{-3}{*}{\shortstack{\\ {EOG}}} 
& ACC & AUC & ACC & AUC & ACC & AUC & ACC & AUC & ACC & AUC & ACC & AUC & ACC & AUC \\
\midrule

$\bullet$ & $\bullet$ & $\bullet$ & 
75.53 & 
89.75 &
\underline{79.92} & 
94.81 & 
75.59 & 
92.82 & 
71.34 & 
90.08 & 
71.49 & 
90.59 & 
78.21 & 
93.14 & 
\textbf{79.64} & 
\textbf{\underline{94.86}} \\

\arrayrulecolor{gray!90}\midrule

$\bullet$ & ~ & ~ &
\begin{tabular}[c]{@{}c@{}}72.42\\[-0.12ex]\scriptsize (-3.11)\end{tabular} &
\begin{tabular}[c]{@{}c@{}}88.57\\[-0.12ex]\scriptsize \underline{(-1.18)}\end{tabular} &
\begin{tabular}[c]{@{}c@{}}74.88\\[-0.12ex]\scriptsize (-5.04)\end{tabular} &
\begin{tabular}[c]{@{}c@{}}90.02\\[-0.12ex]\scriptsize (-4.79)\end{tabular} &
\begin{tabular}[c]{@{}c@{}}73.88\\[-0.12ex]\scriptsize \textcolor{blue}{\underline{(-1.71)}}\end{tabular} &
\begin{tabular}[c]{@{}c@{}}91.50\\[-0.12ex]\scriptsize (-1.32)\end{tabular} &
\begin{tabular}[c]{@{}c@{}}35.87\\[-0.12ex]\scriptsize (-35.47)\end{tabular} &
\begin{tabular}[c]{@{}c@{}}60.93\\[-0.12ex]\scriptsize (-29.15)\end{tabular} &
\begin{tabular}[c]{@{}c@{}}59.42\\[-0.12ex]\scriptsize (-12.07)\end{tabular} &
\begin{tabular}[c]{@{}c@{}}82.23\\[-0.12ex]\scriptsize (-8.36)\end{tabular} &
\begin{tabular}[c]{@{}c@{}}71.72\\[-0.12ex]\scriptsize (-6.49)\end{tabular} &
\begin{tabular}[c]{@{}c@{}}90.35\\[-0.12ex]\scriptsize (-2.79)\end{tabular} &
\begin{tabular}[c]{@{}c@{}}\textbf{\underline{76.31}}\\[-0.12ex]\scriptsize (-3.33)\end{tabular} &
\begin{tabular}[c]{@{}c@{}}\textbf{\underline{93.65}}\\[-0.12ex]\scriptsize \textcolor{blue}{(-1.21)}\end{tabular} \\

\arrayrulecolor{gray!30}\midrule

~ & $\bullet$ & ~ &
\begin{tabular}[c]{@{}c@{}}71.73\\[-0.12ex]\scriptsize \underline{(-3.80)}\end{tabular} &
\begin{tabular}[c]{@{}c@{}}84.82\\[-0.12ex]\scriptsize (-4.93)\end{tabular} &
\begin{tabular}[c]{@{}c@{}}71.76\\[-0.12ex]\scriptsize (-8.16)\end{tabular} &
\begin{tabular}[c]{@{}c@{}}89.13\\[-0.12ex]\scriptsize (-5.68)\end{tabular} &
\begin{tabular}[c]{@{}c@{}}65.92\\[-0.12ex]\scriptsize (-9.67)\end{tabular} &
\begin{tabular}[c]{@{}c@{}}87.95\\[-0.12ex]\scriptsize \textcolor{blue}{\underline{(-4.87)}}\end{tabular} &
\begin{tabular}[c]{@{}c@{}}34.34\\[-0.12ex]\scriptsize (-37.00)\end{tabular} &
\begin{tabular}[c]{@{}c@{}}62.67\\[-0.12ex]\scriptsize (-27.41)\end{tabular} &
\begin{tabular}[c]{@{}c@{}}58.30\\[-0.12ex]\scriptsize (-13.19)\end{tabular} &
\begin{tabular}[c]{@{}c@{}}83.29\\[-0.12ex]\scriptsize (-7.30)\end{tabular} &
\begin{tabular}[c]{@{}c@{}}\underline{\textbf{72.02}}\\[-0.12ex]\scriptsize \textcolor{blue}{(-6.19)}\end{tabular} &
\begin{tabular}[c]{@{}c@{}}88.15\\[-0.12ex]\scriptsize (-4.99)\end{tabular} &
\begin{tabular}[c]{@{}c@{}}71.84\\[-0.12ex]\scriptsize (-7.80)\end{tabular} &
\begin{tabular}[c]{@{}c@{}}\underline{\textbf{89.79}}\\[-0.12ex]\scriptsize (-5.07)\end{tabular} \\

\midrule

~ & ~ & $\bullet$ &
\begin{tabular}[c]{@{}c@{}}72.25\\[-0.12ex]\scriptsize \underline{(-3.28)}\end{tabular} &
\begin{tabular}[c]{@{}c@{}}88.49\\[-0.12ex]\scriptsize \underline{(-1.26)}\end{tabular} &
\begin{tabular}[c]{@{}c@{}}\underline{72.96}\\[-0.12ex]\scriptsize (-6.96)\end{tabular} &
\begin{tabular}[c]{@{}c@{}}\underline{91.99}\\[-0.12ex]\scriptsize (-2.82)\end{tabular} &
\begin{tabular}[c]{@{}c@{}}65.09\\[-0.12ex]\scriptsize (-10.5)\end{tabular} &
\begin{tabular}[c]{@{}c@{}}88.53\\[-0.12ex]\scriptsize (-4.29)\end{tabular} &
\begin{tabular}[c]{@{}c@{}}33.25\\[-0.12ex]\scriptsize (-38.09)\end{tabular} &
\begin{tabular}[c]{@{}c@{}}60.28\\[-0.12ex]\scriptsize (-29.80)\end{tabular} &
\begin{tabular}[c]{@{}c@{}}55.94\\[-0.12ex]\scriptsize (-15.55)\end{tabular} &
\begin{tabular}[c]{@{}c@{}}80.51\\[-0.12ex]\scriptsize (-10.08)\end{tabular} &
\begin{tabular}[c]{@{}c@{}}65.28\\[-0.12ex]\scriptsize (-12.93)\end{tabular} &
\begin{tabular}[c]{@{}c@{}}90.21\\[-0.12ex]\scriptsize \textcolor{blue}{(-2.93)}\end{tabular} &
\begin{tabular}[c]{@{}c@{}}\textbf{72.33}\\[-0.12ex]\scriptsize \textcolor{blue}{(-7.31)}\end{tabular} &
\begin{tabular}[c]{@{}c@{}}\textbf{91.61}\\[-0.12ex]\scriptsize (-3.25)\end{tabular} \\

\midrule

$\bullet$ & $\bullet$ & ~ &
\begin{tabular}[c]{@{}c@{}}75.53\\[-0.12ex]\scriptsize \underline{(-0.00)}\end{tabular} &
\begin{tabular}[c]{@{}c@{}}89.76\\[-0.12ex]\scriptsize \underline{(+0.01)}\end{tabular} &
\begin{tabular}[c]{@{}c@{}}\underline{77.52}\\[-0.12ex]\scriptsize (-2.40)\end{tabular} &
\begin{tabular}[c]{@{}c@{}}\underline{93.57}\\[-0.12ex]\scriptsize (-1.24)\end{tabular} &
\begin{tabular}[c]{@{}c@{}}69.69\\[-0.12ex]\scriptsize (-5.90)\end{tabular} &
\begin{tabular}[c]{@{}c@{}}89.22\\[-0.12ex]\scriptsize (-3.60)\end{tabular} &
\begin{tabular}[c]{@{}c@{}}65.90\\[-0.12ex]\scriptsize (-5.44)\end{tabular} &
\begin{tabular}[c]{@{}c@{}}86.40\\[-0.12ex]\scriptsize (-3.68)\end{tabular} &
\begin{tabular}[c]{@{}c@{}}68.69\\[-0.12ex]\scriptsize (-2.80)\end{tabular} &
\begin{tabular}[c]{@{}c@{}}89.22\\[-0.12ex]\scriptsize (-1.37)\end{tabular} &
\begin{tabular}[c]{@{}c@{}}71.25\\[-0.12ex]\scriptsize (-6.96)\end{tabular} &
\begin{tabular}[c]{@{}c@{}}88.19\\[-0.12ex]\scriptsize (-4.95)\end{tabular} &
\begin{tabular}[c]{@{}c@{}}\textbf{77.08}\\[-0.12ex]\scriptsize \textcolor{blue}{(-2.56)}\end{tabular} &
\begin{tabular}[c]{@{}c@{}}\textbf{93.55}\\[-0.12ex]\scriptsize \textcolor{blue}{(-1.31)}\end{tabular} \\

\midrule

$\bullet$ & ~ & $\bullet$ &
\begin{tabular}[c]{@{}c@{}}75.08\\[-0.12ex]\scriptsize \underline{(-0.45)}\end{tabular} &
\begin{tabular}[c]{@{}c@{}}90.50\\[-0.12ex]\scriptsize \underline{(+0.75)}\end{tabular} &
\begin{tabular}[c]{@{}c@{}}76.44\\[-0.12ex]\scriptsize (-3.48)\end{tabular} &
\begin{tabular}[c]{@{}c@{}}93.46\\[-0.12ex]\scriptsize (-1.35)\end{tabular} &
\begin{tabular}[c]{@{}c@{}}69.00\\[-0.12ex]\scriptsize (-6.59)\end{tabular} &
\begin{tabular}[c]{@{}c@{}}89.91\\[-0.12ex]\scriptsize (-2.91)\end{tabular} &
\begin{tabular}[c]{@{}c@{}}66.11\\[-0.12ex]\scriptsize (-5.23)\end{tabular} &
\begin{tabular}[c]{@{}c@{}}87.73\\[-0.12ex]\scriptsize (-2.35)\end{tabular} &
\begin{tabular}[c]{@{}c@{}}69.00\\[-0.12ex]\scriptsize (-2.49)\end{tabular} &
\begin{tabular}[c]{@{}c@{}}89.91\\[-0.12ex]\scriptsize (-0.68)\end{tabular} &
\begin{tabular}[c]{@{}c@{}}71.44\\[-0.12ex]\scriptsize (-6.77)\end{tabular} &
\begin{tabular}[c]{@{}c@{}}91.01\\[-0.12ex]\scriptsize (-2.13)\end{tabular} &
\begin{tabular}[c]{@{}c@{}}\underline{\textbf{78.34}}\\[-0.12ex]\scriptsize \textcolor{blue}{(-1.30)}\end{tabular} &
\begin{tabular}[c]{@{}c@{}}\underline{\textbf{94.47}}\\[-0.12ex]\scriptsize \textcolor{blue}{(-0.39)}\end{tabular} \\

\midrule

~ & $\bullet$ & $\bullet$ &
\begin{tabular}[c]{@{}c@{}}75.12\\[-0.12ex]\scriptsize \underline{(-0.41)}\end{tabular} &
\begin{tabular}[c]{@{}c@{}}89.63\\[-0.12ex]\scriptsize \underline{(-0.12)}\end{tabular} &
\begin{tabular}[c]{@{}c@{}}75.30\\[-0.12ex]\scriptsize (-4.62)\end{tabular} &
\begin{tabular}[c]{@{}c@{}}93.00\\[-0.12ex]\scriptsize (-1.81)\end{tabular} &
\begin{tabular}[c]{@{}c@{}}70.27\\[-0.12ex]\scriptsize (-5.32)\end{tabular} &
\begin{tabular}[c]{@{}c@{}}89.86\\[-0.12ex]\scriptsize (-2.96)\end{tabular} &
\begin{tabular}[c]{@{}c@{}}58.39\\[-0.12ex]\scriptsize (-12.95)\end{tabular} &
\begin{tabular}[c]{@{}c@{}}80.36\\[-0.12ex]\scriptsize (-9.72)\end{tabular} &
\begin{tabular}[c]{@{}c@{}}70.27\\[-0.12ex]\scriptsize \textcolor{blue}{(-1.22)}\end{tabular} &
\begin{tabular}[c]{@{}c@{}}87.84\\[-0.12ex]\scriptsize (-2.75)\end{tabular} &
\begin{tabular}[c]{@{}c@{}}72.72\\[-0.12ex]\scriptsize (-5.49)\end{tabular} &
\begin{tabular}[c]{@{}c@{}}91.68\\[-0.12ex]\scriptsize (-1.46)\end{tabular} &
\begin{tabular}[c]{@{}c@{}}\underline{\textbf{78.10}}\\[-0.12ex]\scriptsize (-1.54)\end{tabular} &
\begin{tabular}[c]{@{}c@{}}\underline{\textbf{93.87}}\\[-0.12ex]\scriptsize \textcolor{blue}{(-0.99)}\end{tabular} \\

\arrayrulecolor{black} % 이후 선 다시 검정
\midrule
\multicolumn{3}{c}{\textit{Mean Absolute Value (MAV)}} &  
\begin{tabular}[c]{@{}c@{}}73.69\\[-0.12ex]\scriptsize \underline{(1.84)}\end{tabular} &
\begin{tabular}[c]{@{}c@{}}88.63\\[-0.12ex]\scriptsize \underline{(1.38)}\end{tabular} &
\begin{tabular}[c]{@{}c@{}}74.81\\[-0.12ex]\scriptsize (5.11)\end{tabular} &
\begin{tabular}[c]{@{}c@{}}91.86\\[-0.12ex]\scriptsize (2.95)\end{tabular} &
\begin{tabular}[c]{@{}c@{}}68.98\\[-0.12ex]\scriptsize (6.62)\end{tabular} &
\begin{tabular}[c]{@{}c@{}}89.50\\[-0.12ex]\scriptsize (3.33)\end{tabular} &
\begin{tabular}[c]{@{}c@{}}48.98\\[-0.12ex]\scriptsize (22.36)\end{tabular} &
\begin{tabular}[c]{@{}c@{}}73.06\\[-0.12ex]\scriptsize (17.02)\end{tabular} &
\begin{tabular}[c]{@{}c@{}}63.60\\[-0.12ex]\scriptsize (7.89)\end{tabular} &
\begin{tabular}[c]{@{}c@{}}85.50\\[-0.12ex]\scriptsize (5.09)\end{tabular} &
\begin{tabular}[c]{@{}c@{}}70.74\\[-0.12ex]\scriptsize (7.47)\end{tabular} &
\begin{tabular}[c]{@{}c@{}}89.93\\[-0.12ex]\scriptsize (3.21)\end{tabular} &
\begin{tabular}[c]{@{}c@{}}\underline{\textbf{75.67}}\\[-0.12ex]\scriptsize \textcolor{blue}{(3.97)}\end{tabular} &
\begin{tabular}[c]{@{}c@{}}\underline{\textbf{92.82}}\\[-0.12ex]\scriptsize \textcolor{blue}{(2.04)}\end{tabular} \\

\bottomrule

\end{tabular}
\caption{Comparison with other methodologies for sleep stage classification using Sleep-EDFX.}
\label{table:table_1}
\end{table*}
%%%%%%%%%%%%%%%%%%%%%%%%%%%%%%%%%%%%%%%%%%%%%%%%%%%%%%%%%%%%%%%%%%%%%%%%%%%%%%%%%%%%%%%%%%%%%%%%%%%%%%%%%%%%%%%%%%%%%%%%%%%%%%%%%%%%%%%%%

%%%%%%%%%%%%%%%%%%%%%%%%%%%%%%%%%%%%%%%%%%%%%%%%%%%%%%%%%%%%%%%%%%%%%%%%%%%%%%%%%%%%%%%%%%%%%%%%%%%%%%%%%%%%%%%%%%%%%%%%%%%%%%%%%%%%%%%%%
% Table 2

\begin{table*}[!t]
\centering
\footnotesize
\def\arraystretch{0.8}
\setlength{\tabcolsep}{3.2pt}
\begin{tabular}{
>{\centering\arraybackslash}m{1.2cm} 
>{\centering\arraybackslash}m{1.2cm} 
>{\centering\arraybackslash}m{1.0cm} 
cc cc cc cc cc cc cc}
\toprule
\multicolumn{3}{c}{\textit{Modality}} & \multicolumn{4}{c}{\textit{Dedicated Model}} & \multicolumn{10}{c}{\textit{Single Model}} \\
\cmidrule(lr){1-3} \cmidrule(lr){4-7} \cmidrule(lr){8-17}
& & 
& \multicolumn{2}{c}{ContraWR} & \multicolumn{2}{c}{SynthSleepNet} 
& \multicolumn{2}{c}{MultiMAE} & \multicolumn{2}{c}{RobustSsF} 
& \multicolumn{2}{c}{CroSSL} & \multicolumn{2}{c}{MaskMentor} 
& \multicolumn{2}{c}{PhysioME} \\
\cmidrule(lr){4-5} \cmidrule(lr){6-7} \cmidrule(lr){8-9}
\cmidrule(lr){10-11} \cmidrule(lr){12-13} \cmidrule(lr){14-15} \cmidrule(lr){16-17}
\multirow{-3}{*}{\shortstack{\\ {ABP}}} 
& \multirow{-3}{*}{\shortstack{\\ {ECG}}} 
& \multirow{-3}{*}{\shortstack{\\ {PPG}}} 
& ACC & AUC & ACC & AUC & ACC & AUC & ACC & AUC & ACC & AUC & ACC & AUC & ACC & AUC \\
\midrule

$\bullet$ & $\bullet$ & $\bullet$ & 
91.53 & 
87.04 & 
89.36 & 
\underline{94.97} & 
84.21 & 
79.79 & 
85.19 & 
61.75 & 
80.17 & 
67.28 & 
\underline{\textbf{92.52}} & 
85.49 & 
89.59 & 
\textbf{94.71} \\

\arrayrulecolor{gray!90}\midrule

$\bullet$ & ~ & ~ &
\begin{tabular}[c]{@{}c@{}}82.13\\[-0.12ex]\scriptsize (-9.40)\end{tabular} &
\begin{tabular}[c]{@{}c@{}}90.35\\[-0.12ex]\scriptsize \underline{(+3.31)}\end{tabular} &
\begin{tabular}[c]{@{}c@{}}\underline{89.30}\\[-0.12ex]\scriptsize \underline{(-0.06)}\end{tabular} &
\begin{tabular}[c]{@{}c@{}}\underline{94.81}\\[-0.12ex]\scriptsize (-0.16)\end{tabular} &
\begin{tabular}[c]{@{}c@{}}66.14\\[-0.12ex]\scriptsize (-18.07)\end{tabular} &
\begin{tabular}[c]{@{}c@{}}73.49\\[-0.12ex]\scriptsize (-6.30)\end{tabular} &
\begin{tabular}[c]{@{}c@{}}62.55\\[-0.12ex]\scriptsize (-22.64)\end{tabular} &
\begin{tabular}[c]{@{}c@{}}53.88\\[-0.12ex]\scriptsize (-7.87)\end{tabular} &
\begin{tabular}[c]{@{}c@{}}72.43\\[-0.12ex]\scriptsize (-7.74)\end{tabular} &
\begin{tabular}[c]{@{}c@{}}51.50\\[-0.12ex]\scriptsize (-15.78)\end{tabular} &
\begin{tabular}[c]{@{}c@{}}82.90\\[-0.12ex]\scriptsize (-9.62)\end{tabular} &
\begin{tabular}[c]{@{}c@{}}60.01\\[-0.12ex]\scriptsize (-25.48)\end{tabular} &
\begin{tabular}[c]{@{}c@{}}\textbf{88.90}\\[-0.12ex]\scriptsize \textcolor{blue}{(-0.69)}\end{tabular} &
\begin{tabular}[c]{@{}c@{}}\textbf{94.38}\\[-0.12ex]\scriptsize \textcolor{blue}{(-0.33)}\end{tabular} \\

\arrayrulecolor{gray!30}\midrule

~ & $\bullet$ & ~ &
\begin{tabular}[c]{@{}c@{}}85.61\\[-0.12ex]\scriptsize (-5.92)\end{tabular} &
\begin{tabular}[c]{@{}c@{}}83.17\\[-0.12ex]\scriptsize (-3.87)\end{tabular} &
\begin{tabular}[c]{@{}c@{}}\underline{86.97}\\[-0.12ex]\scriptsize \underline{(-2.39)}\end{tabular} &
\begin{tabular}[c]{@{}c@{}}\underline{94.84}\\[-0.12ex]\scriptsize \underline{(-0.13)}\end{tabular} &
\begin{tabular}[c]{@{}c@{}}71.70\\[-0.12ex]\scriptsize (-12.51)\end{tabular} &
\begin{tabular}[c]{@{}c@{}}52.52\\[-0.12ex]\scriptsize (-27.27)\end{tabular} &
\begin{tabular}[c]{@{}c@{}}75.56\\[-0.12ex]\scriptsize (-9.63)\end{tabular} &
\begin{tabular}[c]{@{}c@{}}53.50\\[-0.12ex]\scriptsize \textcolor{blue}{(-8.25)}\end{tabular} &
\begin{tabular}[c]{@{}c@{}}71.86\\[-0.12ex]\scriptsize \textcolor{blue}{(-8.31)}\end{tabular} &
\begin{tabular}[c]{@{}c@{}}49.98\\[-0.12ex]\scriptsize (-17.30)\end{tabular} &
\begin{tabular}[c]{@{}c@{}}\textbf{81.73}\\[-0.12ex]\scriptsize (-10.79)\end{tabular} &
\begin{tabular}[c]{@{}c@{}}59.85\\[-0.12ex]\scriptsize (-25.64)\end{tabular} &
\begin{tabular}[c]{@{}c@{}}78.54\\[-0.12ex]\scriptsize (-11.05)\end{tabular} &
\begin{tabular}[c]{@{}c@{}}\textbf{77.80}\\[-0.12ex]\scriptsize (-16.91)\end{tabular} \\

\midrule

~ & ~ & $\bullet$ &
\begin{tabular}[c]{@{}c@{}}81.34\\[-0.12ex]\scriptsize (-10.19)\end{tabular} &
\begin{tabular}[c]{@{}c@{}}74.52\\[-0.12ex]\scriptsize (-12.52)\end{tabular} &
\begin{tabular}[c]{@{}c@{}}\underline{87.65}\\[-0.12ex]\scriptsize (-1.71)\end{tabular} &
\begin{tabular}[c]{@{}c@{}}\underline{94.58}\\[-0.12ex]\scriptsize \underline{(-0.39)}\end{tabular} &
\begin{tabular}[c]{@{}c@{}}\textbf{87.62}\\[-0.12ex]\scriptsize \textcolor{blue}{\underline{(+3.41)}}\end{tabular} &
\begin{tabular}[c]{@{}c@{}}68.28\\[-0.12ex]\scriptsize (-11.51)\end{tabular} &
\begin{tabular}[c]{@{}c@{}}77.58\\[-0.12ex]\scriptsize (-7.61)\end{tabular} &
\begin{tabular}[c]{@{}c@{}}53.99\\[-0.12ex]\scriptsize \textcolor{blue}{(-7.76)}\end{tabular} &
\begin{tabular}[c]{@{}c@{}}71.49\\[-0.12ex]\scriptsize (-8.68)\end{tabular} &
\begin{tabular}[c]{@{}c@{}}49.17\\[-0.12ex]\scriptsize (-18.11)\end{tabular} &
\begin{tabular}[c]{@{}c@{}}82.70\\[-0.12ex]\scriptsize (-9.82)\end{tabular} &
\begin{tabular}[c]{@{}c@{}}59.72\\[-0.12ex]\scriptsize (-25.77)\end{tabular} &
\begin{tabular}[c]{@{}c@{}}81.14\\[-0.12ex]\scriptsize (-8.45)\end{tabular} &
\begin{tabular}[c]{@{}c@{}}\textbf{79.92}\\[-0.12ex]\scriptsize (-14.79)\end{tabular} \\

\midrule

$\bullet$ & $\bullet$ & ~ &
\begin{tabular}[c]{@{}c@{}}89.02\\[-0.12ex]\scriptsize (-2.51)\end{tabular} &
\begin{tabular}[c]{@{}c@{}}90.16\\[-0.12ex]\scriptsize \underline{(+3.12)}\end{tabular} &
\begin{tabular}[c]{@{}c@{}}\underline{89.65}\\[-0.12ex]\scriptsize \underline{(+0.29)}\end{tabular} &
\begin{tabular}[c]{@{}c@{}}94.04\\[-0.12ex]\scriptsize (-0.93)\end{tabular} &
\begin{tabular}[c]{@{}c@{}}72.23\\[-0.12ex]\scriptsize (-11.98)\end{tabular} &
\begin{tabular}[c]{@{}c@{}}60.47\\[-0.12ex]\scriptsize (-19.32)\end{tabular} &
\begin{tabular}[c]{@{}c@{}}77.51\\[-0.12ex]\scriptsize (-7.68)\end{tabular} &
\begin{tabular}[c]{@{}c@{}}55.07\\[-0.12ex]\scriptsize (-6.68)\end{tabular} &
\begin{tabular}[c]{@{}c@{}}71.25\\[-0.12ex]\scriptsize (-8.92)\end{tabular} &
\begin{tabular}[c]{@{}c@{}}51.11\\[-0.12ex]\scriptsize (-16.17)\end{tabular} &
\begin{tabular}[c]{@{}c@{}}83.69\\[-0.12ex]\scriptsize (-8.83)\end{tabular} &
\begin{tabular}[c]{@{}c@{}}63.28\\[-0.12ex]\scriptsize (-22.21)\end{tabular} &
\begin{tabular}[c]{@{}c@{}}\textbf{89.40}\\[-0.12ex]\scriptsize \textcolor{blue}{(-0.19)}\end{tabular} &
\begin{tabular}[c]{@{}c@{}}\underline{\textbf{94.63}}\\[-0.12ex]\scriptsize \textcolor{blue}{(-0.08)}\end{tabular} \\

\midrule

$\bullet$ & ~ & $\bullet$ &
\begin{tabular}[c]{@{}c@{}}89.36\\[-0.12ex]\scriptsize (-2.17)\end{tabular} &
\begin{tabular}[c]{@{}c@{}}92.96\\[-0.12ex]\scriptsize \underline{(+5.92)}\end{tabular} &
\begin{tabular}[c]{@{}c@{}}89.07\\[-0.12ex]\scriptsize (-0.29)\end{tabular} &
\begin{tabular}[c]{@{}c@{}}\underline{94.98}\\[-0.12ex]\scriptsize (+0.01)\end{tabular} &
\begin{tabular}[c]{@{}c@{}}79.68\\[-0.12ex]\scriptsize (-4.53)\end{tabular} &
\begin{tabular}[c]{@{}c@{}}72.63\\[-0.12ex]\scriptsize (-7.16)\end{tabular} &
\begin{tabular}[c]{@{}c@{}}78.53\\[-0.12ex]\scriptsize (-6.66)\end{tabular} &
\begin{tabular}[c]{@{}c@{}}55.86\\[-0.12ex]\scriptsize (-5.89)\end{tabular} &
\begin{tabular}[c]{@{}c@{}}75.85\\[-0.12ex]\scriptsize (-4.32)\end{tabular} &
\begin{tabular}[c]{@{}c@{}}58.88\\[-0.12ex]\scriptsize (-8.40)\end{tabular} &
\begin{tabular}[c]{@{}c@{}}84.11\\[-0.12ex]\scriptsize (-8.41)\end{tabular} &
\begin{tabular}[c]{@{}c@{}}63.86\\[-0.12ex]\scriptsize (-21.63)\end{tabular} &
\begin{tabular}[c]{@{}c@{}}\underline{\textbf{89.38}}\\[-0.12ex]\scriptsize \textcolor{blue}{\underline{(-0.21)}}\end{tabular} &
\begin{tabular}[c]{@{}c@{}}\textbf{94.36}\\[-0.12ex]\scriptsize \textcolor{blue}{(-0.35)}\end{tabular} \\

\midrule

~ & $\bullet$ & $\bullet$ &
\begin{tabular}[c]{@{}c@{}}88.34\\[-0.12ex]\scriptsize (-3.19)\end{tabular} &
\begin{tabular}[c]{@{}c@{}}77.46\\[-0.12ex]\scriptsize (-9.58)\end{tabular} &
\begin{tabular}[c]{@{}c@{}}86.81\\[-0.12ex]\scriptsize (-2.55)\end{tabular} &
\begin{tabular}[c]{@{}c@{}}\underline{94.84}\\[-0.12ex]\scriptsize (-0.13)\end{tabular} &
\begin{tabular}[c]{@{}c@{}}83.10\\[-0.12ex]\scriptsize (-1.11)\end{tabular} &
\begin{tabular}[c]{@{}c@{}}61.59\\[-0.12ex]\scriptsize (-18.20)\end{tabular} &
\begin{tabular}[c]{@{}c@{}}79.01\\[-0.12ex]\scriptsize (-6.18)\end{tabular} &
\begin{tabular}[c]{@{}c@{}}56.55\\[-0.12ex]\scriptsize (-5.20)\end{tabular} &
\begin{tabular}[c]{@{}c@{}}74.10\\[-0.12ex]\scriptsize (-6.07)\end{tabular} &
\begin{tabular}[c]{@{}c@{}}56.49\\[-0.12ex]\scriptsize (-10.79)\end{tabular} &
\begin{tabular}[c]{@{}c@{}}88.39\\[-0.12ex]\scriptsize (-4.13)\end{tabular} &
\begin{tabular}[c]{@{}c@{}}74.92\\[-0.12ex]\scriptsize (-10.57)\end{tabular} &
\begin{tabular}[c]{@{}c@{}}\underline{\textbf{88.96}}\\[-0.12ex]\scriptsize \textcolor{blue}{\underline{(-0.63)}}\end{tabular} &
\begin{tabular}[c]{@{}c@{}}\textbf{94.59}\\[-0.12ex]\scriptsize \textcolor{blue}{\underline{(-0.12)}}\end{tabular} \\

\arrayrulecolor{black} % 이후 선 다시 검정
\midrule
\multicolumn{3}{c}{\textit{Mean Absolute Value (MAV)}} &  
\begin{tabular}[c]{@{}c@{}}85.97\\[-0.12ex]\scriptsize (5.56)\end{tabular} &
\begin{tabular}[c]{@{}c@{}}84.77\\[-0.12ex]\scriptsize (6.39)\end{tabular} &
\begin{tabular}[c]{@{}c@{}}\underline{88.24}\\[-0.12ex]\scriptsize \underline{(1.22)}\end{tabular} &
\begin{tabular}[c]{@{}c@{}}\underline{94.68}\\[-0.12ex]\scriptsize \underline{(0.29)}\end{tabular} &
\begin{tabular}[c]{@{}c@{}}76.75\\[-0.12ex]\scriptsize (8.60)\end{tabular} &
\begin{tabular}[c]{@{}c@{}}64.83\\[-0.12ex]\scriptsize (14.96)\end{tabular} &
\begin{tabular}[c]{@{}c@{}}75.12\\[-0.12ex]\scriptsize (10.07)\end{tabular} &
\begin{tabular}[c]{@{}c@{}}54.81\\[-0.12ex]\scriptsize (6.94)\end{tabular} &
\begin{tabular}[c]{@{}c@{}}72.83\\[-0.12ex]\scriptsize (7.34)\end{tabular} &
\begin{tabular}[c]{@{}c@{}}52.86\\[-0.12ex]\scriptsize (14.43)\end{tabular} &
\begin{tabular}[c]{@{}c@{}}83.92\\[-0.12ex]\scriptsize (8.60)\end{tabular} &
\begin{tabular}[c]{@{}c@{}}63.61\\[-0.12ex]\scriptsize (21.88)\end{tabular} &
\begin{tabular}[c]{@{}c@{}}\textbf{86.05}\\[-0.12ex]\scriptsize \textcolor{blue}{(3.54)}\end{tabular} &
\begin{tabular}[c]{@{}c@{}}\textbf{89.28}\\[-0.12ex]\scriptsize \textcolor{blue}{(5.43)}\end{tabular} \\
\bottomrule
\end{tabular}
\begin{flushleft}
\scriptsize
\begin{itemize}
  \setlength\itemsep{-0.3em}
  \item The numbers in parentheses indicate changes in performance compared to the full-modality scenario, except for the last row.
  \item The MAV of each performance and the changes in performance are shown in the last row.
  \item The best results within single-models in each row are shown in bold.
  \item The best results in each row are underlined. 
  \item  The smallest change in performance within single-models in each row is shown in blue, compared with the full-modality scenario.
\end{itemize}
\end{flushleft}

\caption{Comparison with other methodologies for hypotension prediction using VitalDB.}
\label{table:table_2}
\end{table*}
%%%%%%%%%%%%%%%%%%%%%%%%%%%%%%%%%%%%%%%%%%%%%%%%%%%%%%%%%%%%%%%%%%%%%%%%%%%%%%%%%%%%%%%%%%%%%%%%%%%%%%%%%%%%%%%%%%%%%%%%%%%%%%%%%%%%%%%%%

\section{Experiments} 
\subsection{Dataset Description} \subsubsection{Sleep-EDFX \cite{kemp2000analysis}} contains polysomnographic (PSG) recordings and aims to classify data into five sleep stages based on the American Academy of Sleep Medicine guidelines. This study utilized the SC subset and three channels: electroencephalography (EEG) Fpz-Cz, EEG Pz-Oz, and horizontal electrooculography (EOG). A 0-40 Hz band-pass filter was applied to all signals.

\subsubsection{VitalDB \cite{lee2022vitaldb}} comprises intraoperative physiological recordings from 6,388 surgical cases, including 486,451 waveform segments and the corresponding perioperative patient information. This study aimed to predict intraoperative hypotension within the following 5 min using three types of data: arterial blood pressure (ABP), electrocardiography (ECG), and photoplethysmography (PPG) signals.

\subsection{Baselines} 
Four multimodal SSL frameworks for missing modality were implemented as baselines. Except for CroSSL, the selected models were originally developed for vision tasks; thus, the signals were converted into short-time Fourier transform images and used as inputs for training. The baselines were as follows:

\begin{itemize}
    \item \textbf{MultiMAE \cite{bachmann2022multimae}} is an MAE-based approach that learns joint representations by reconstructing masked inputs across multiple modalities.
    \item \textbf{RobustSsF \cite{lee2023robustssf}} employs four independent encoder-decoder branches to separately process each modality. To preserve inter-modality consistency, this model adopts an SSL scheme based on coupled regularization loss.
    \item \textbf{CroSSL \cite{deldari2024crossl}} performs masking in the latent space and learns global representations from heterogeneous sensor modalities.
    \item \textbf{MaskMentor \cite{zhao2024maskmentor}} is a masked self-teaching framework in which a teacher model is trained using complete multimodal inputs, whereas a student model learns from partially missing inputs, using the predictions of the teacher as pseudo-labels.
\end{itemize}

\subsection{Evaluation Scheme} To evaluate the performance of each method, 5-fold subject-group cross-validation was conducted. The datasets were divided into three subsets: pretrain, train, and test. The pretrain subset was utilized for SSL without labels and the train subset was used for linear evaluation with a limited amount of labeled data. A downstream classifier was attached to the pretrained network to perform two downstream tasks. Finally, the performance of the proposed model and baselines was investigated using the test subset.

\subsection{Evaluation Metric} Model performance was evaluated using accuracy ($\mathrm{ACC}$) and the area under the receiver operating characteristic (ROC) curve ($\mathrm{AUC}$). $\mathrm{AUC}$ is a reliable evaluation metric for tasks with class imbalance, such as sleep stage classification and hypotension prediction.

\subsubsection{ACC} indicates the proportion of correctly classified samples and provides an overall assessment of model performance across the entire dataset. $\mathrm{ACC}$ is defined as:

\begin{equation}
\mathrm{ACC} = \frac{\mathrm{TP} + \mathrm{TN}}{\mathrm{TP} + \mathrm{TN} + \mathrm{FP} + \mathrm{FN}}
\end{equation} where $\mathrm{TP}$ denotes true positives, $\mathrm{TN}$ denotes true negatives, $\mathrm{FP}$ denotes false positives, and $\mathrm{FN}$ denotes false negatives.

\subsubsection{AUC} quantifies the area under the ROC curve, which plots the true positive rate ($\mathrm{TPR}$) against the false positive rate ($\mathrm{FPR}$). The ROC curve plots the $\mathrm{TPR}$ on the $y$-axis against the FPR on the $x$-axis. $\mathrm{AUC}$ is defined as follows:

\begin{equation}
\mathrm{AUC} = \int_{0}^{1} \mathrm{TPR}(\mathrm{FPR})\, d\mathrm{FPR}
\end{equation}
where $\mathrm{TPR}$ and $\mathrm{FPR}$ are calculated as follows: 
$\mathrm{TPR} = \frac{\mathrm{TP}}{\mathrm{TP} + \mathrm{FN}}$ and 
$\mathrm{FPR} = \frac{\mathrm{FP}}{\mathrm{FP} + \mathrm{TN}}$.

\section{Results}
\subsection{Comparison of Other Methodologies}
Two downstream tasks were conducted to evaluate the proposed model: (1) sleep stage classification using the Sleep-EDFX dataset (Table \ref{table:table_1}), and (2) hypotension prediction using the VitalDB dataset (Table \ref{table:table_2}). Its performance was compared with those of existing single-model approaches (i.e., MultiMAE \cite{bachmann2022multimae}, RobustSsF \cite{lee2023robustssf}, CroSSL \cite{deldari2024crossl}, and MaskMentor \cite{zhao2024maskmentor}) and dedicated-model approaches (i.e., ContraWR \cite{yang2021self} and SynthSleepNet \cite{lee2025toward}). In addition, various modality combination experiments were performed to assess the performance and robustness of the model under missing modality scenarios.

\subsubsection{Comparison within Single-Model Approaches} ~ 

\paragraph{\textit{(Performance under Missing Modalities)}} PhysioME consistently outperformed the existing methods across a wide range of missing modality scenarios. For both downstream tasks, it achieved the highest average $\mathrm{ACC}$ and $\mathrm{AUC}$ across all the missing modality combinations. In the sleep stage classification task, PhysioME demonstrated the best performance in nearly all scenarios, except for the single-modality case using EEG Pz-Cz (bold highlights in Table 1). In the hypotension prediction task, although some models showed slightly better $\mathrm{ACC}$ in specific cases (e.g., single-modality ECG and PPG), PhysioME achieved the highest $\mathrm{AUC}$ in all scenarios (bold highlights in Table 2).

\paragraph{\textit{(Robustness under Missing Modalities)}} PhysioME exhibited the smallest performance degradation compared to the full-modality setting in most missing modality scenarios, maintaining stable accuracy and $\mathrm{AUC}$ across conditions (blue highlights in Tables 1 and 2). When comparing the mean absolute difference in performance relative to the full-modality case, PhysioME showed the lowest values across all metrics: 3.97\% ($\mathrm{ACC}$) and 2.04 ($\mathrm{AUC}$) for sleep stage classification, and 3.54\% ($\mathrm{ACC}$) and 5.43 ($\mathrm{AUC}$) for hypotension prediction. These results indicate that PhysioME maintained stable performance even under missing modality conditions.

\subsubsection{Comparison between Single-Model and Dedicated-Model Approaches} ~ 

\paragraph{\textit{(Competitiveness Against Task-Specific Models)}} Despite being a single-model framework, PhysioME demonstrated performance comparable to or better than dedicated models across various modality combinations for both downstream tasks (underlines in Tables \ref{table:table_1} and \ref{table:table_2}). This challenges the common assumption that dedicated models are inherently more advantageous, indicating that a single-model can achieve a robust and reliable performance.

Specifically, in the sleep stage classification task, PhysioME consistently outperformed ContraWR across all modality combinations and surpassed SynthSleepNet in most cases. In the hypotension prediction task, PhysioME achieved a performance on par with that of the dedicated models. Notably, in terms of the $\mathrm{AUC}$, PhysioME outperformed ContraWR in all scenarios except for the single-modality ECG case. While SynthSleepNet exhibited a higher overall $\mathrm{AUC}$, PhysioME achieved superior results for the ABP+PPG and ECG+PPG combinations.

\subsection{Ablation Studies}
Ablation studies were conducted on the Sleep-EDFX dataset to determine the optimal configuration for PhysioME. The performance was evaluated under three modality settings: (i) EEG Fpz-Cz, (ii) EEG Fpz-Cz + EOG, and (iii) EEG Fpz-Cz + EEG Pz-Cz + EOG, while being limited to 10 epochs for training efficiency.

\subsubsection{Modality Backbone Network} 
As listed in Table \ref{table:table_3}, the DP-NeuroNet backbone achieved the highest overall performance, confirming its superior ability to encode the distinctive characteristics of physiological signals. Specifically, CNN-based architectures (e.g., ResNet and EfficientNet) outperformed ViT-based architectures (e.g., ViT and Swin Transformer). Moreover, DP-NeuroNet surpassed the original NeuroNet across all modality combinations; in the single-modality setting, it improved $\mathrm{ACC}$ by 6.09\% and $\mathrm{AUC}$ by 3.22.

%%%%%%%%%%%%%%%%%%%%%%%%%%%%%%%%%%%%%%%%%%%%%%%%%%%%%%%%%%%%%%%%%%%%%%%%%%%%%%%%%%%%%%%%%%%%%%%%%%%%%%%%%%%%%%%%%%%%%%%%%%%%%%%%%%%%%%%%%
% Table 3

\begin{table}[htbp]

\centering
\scriptsize
\def\arraystretch{0.75}
\begin{tabular}{c  cc  cc  cc}
\toprule
\multirow{2}{*}{\shortstack{\\ {\textit{Model Name}}}} 
& \multicolumn{2}{c}{\textit{Modality 1}} 
& \multicolumn{2}{c}{\textit{Modality 2}} 
& \multicolumn{2}{c}{\textit{Full}} \\
\cmidrule(lr){2-3} \cmidrule(lr){4-5} \cmidrule(lr){6-7}
& ACC & AUC & ACC & AUC & ACC & AUC \\
\midrule
ResNet           & 70.13 & 89.12 & 71.69 & 89.80 & 74.66 & 90.97 \\
EfficientNet     & 68.66 & 86.21 & 72.01 & 88.54 & 74.38 & 89.53 \\
ViT              & 42.68 & 66.08 & 47.16 & 66.99 & 48.18 & 65.28 \\
Swin Transformer & 51.22 & 73.76 & 55.95 & 75.79 & 57.19 & 75.65 \\
NeuroNet         & 70.42 & 89.48 & 74.43 & 91.23 & 75.90 & 92.02 \\
\textbf{DP-NeuroNet} 
                 & \textbf{76.51} & \textbf{92.70} 
                 & \textbf{77.77} & \textbf{93.85} 
                 & \textbf{78.83} & \textbf{94.46} \\
\bottomrule
\end{tabular}
\caption{Comparison of modality backbone networks.}
\label{table:table_3}
\end{table}

%%%%%%%%%%%%%%%%%%%%%%%%%%%%%%%%%%%%%%%%%%%%%%%%%%%%%%%%%%%%%%%%%%%%%%%%%%%%%%%%%%%%%%%%%%%%%%%%%%%%%%%%%%%%%%%%%%%%%%%%%%%%%%%%%%%%%%%%%

\subsubsection{Presence or Absence of a Gradient on Restoration Decoder} 
This study investigated whether the gradients from the restoration decoder should be back-propagated to the upstream network components. As listed in Table \ref{table:table_4}, disabling the gradient propagation (i.e., applying a stop-gradient operation) consistently outperformed the alternative across all modality combinations. These findings suggest that isolating the restoration decoder enables it to focus more effectively on reconstructing missing modality tokens, thereby yielding superior overall performance. 

%%%%%%%%%%%%%%%%%%%%%%%%%%%%%%%%%%%%%%%%%%%%%%%%%%%%%%%%%%%%%%%%%%%%%%%%%%%%%%%%%%%%%%%%%%%%%%%%%%%%%%%%%%%%%%%%%%%%%%%%%%%%%%%%%%%%%%%%%%
% Table 4
\begin{table}[htbp]
\centering
\scriptsize
\def\arraystretch{0.75}
\begin{tabular}{c  cc  cc  cc}
\toprule
\multirow{2}{*}{\shortstack{\\ {\textit{Gradient P/A}}}} 
& \multicolumn{2}{c}{\textit{Modality 1}} 
& \multicolumn{2}{c}{\textit{Modality 2}} 
& \multicolumn{2}{c}{\textit{Full}} \\
\cmidrule(lr){2-3} \cmidrule(lr){4-5} \cmidrule(lr){6-7}
& ACC & AUC & ACC & AUC & ACC & AUC \\
\midrule
with gradient       & 75.54 & 92.31 & 76.52 & 93.21 & 77.13 & 93.38 \\
\textbf{without gradient} 
                    & \textbf{76.51} & \textbf{92.70} 
                    & \textbf{77.77} & \textbf{93.85} 
                    & \textbf{78.83} & \textbf{94.46} \\
\bottomrule
\end{tabular}
\begin{flushleft}
\scriptsize * P/A = Presence or Absence.
\end{flushleft}

\caption{Performance based on the presence or absence of a gradient.}
\label{table:table_4}
\end{table}
%%%%%%%%%%%%%%%%%%%%%%%%%%%%%%%%%%%%%%%%%%%%%%%%%%%%%%%%%%%%%%%%%%%%%%%%%%%%%%%%%%%%%%%%%%%%%%%%%%%%%%%%%%%%%%%%%%%%%%%%%%%%%%%%%%%%%%%%%

\subsubsection{Restoration Strategy} 
To determine the optimal strategy for handling missing modalities in PhysioME, three alternatives were evaluated: (i) masked tokens, (ii) memory tokens, and (iii) restoration decoder tokens (Table \ref{table:table_5}). In previous studies \cite{woo2023towards}, memory tokens have been variously denoted as modality or learnable tokens. The results of this study show that the proposed restoration decoder approach delivered the highest performance across all metrics and modality configurations. Notably, in the single-modality scenario, the restoration decoder strategy outperformed the masked and memory token strategies by 4.94\% and 2.29\% in $\mathrm{ACC}$, respectively.

%%%%%%%%%%%%%%%%%%%%%%%%%%%%%%%%%%%%%%%%%%%%%%%%%%%%%%%%%%%%%%%%%%%%%%%%%%%%%%%%%%%%%%%%%%%%%%%%%%%%%%%%%%%%%%%%%%%%%%%%%%%%%%%%%%%%%%%%%
% Table 5

\begin{table}[htbp]

\centering
\scriptsize
\def\arraystretch{0.75}
\begin{tabular}{c  cc  cc  cc}
\toprule
\multirow{2}{*}{\shortstack{\\ \textit{Restoration} \\ \textit{Strategy}}} 
& \multicolumn{2}{c}{\textit{Modality 1}} 
& \multicolumn{2}{c}{\textit{Modality 2}} 
& \multicolumn{2}{c}{\textit{Full}} \\
\cmidrule(lr){2-3} \cmidrule(lr){4-5} \cmidrule(lr){6-7}
& ACC & AUC & ACC & AUC & ACC & AUC \\
\midrule
Masked Token        & 71.57 & 85.47 & 73.02 & 87.16 & 74.50 & 87.63 \\
Memory Token        & 74.22 & 91.63 & 75.79 & 91.80 & 77.08 & 93.43 \\
\textbf{Ours}       
                    & \textbf{76.51} & \textbf{92.70} 
                    & \textbf{77.77} & \textbf{93.85} 
                    & \textbf{78.83} & \textbf{94.46} \\
\bottomrule
\end{tabular}
\begin{flushleft}
\scriptsize {* Ours: Restoration Decoder based Token}
\end{flushleft}

\caption{Performance of the restoration strategy based on PhysioME.}
\label{table:table_5}
\end{table}
%%%%%%%%%%%%%%%%%%%%%%%%%%%%%%%%%%%%%%%%%%%%%%%%%%%%%%%%%%%%%%%%%%%%%%%%%%%%%%%%%%%%%%%%%%%%%%%%%%%%%%%%%%%%%%%%%%%%%%%%%%%%%%%%%%%%%%%%%

\subsubsection{Restoration Decoder Dimension and Depth} 

Table \ref{table:table_6} lists the effects of the decoder dimensions and depth on the model performance. Overall, increasing either parameter led to systematic improvements in all evaluation metrics. In particular, a configuration with 512 hidden dimensions and a depth of eight layers yielded the best results for most modality combinations. These findings suggest that a high-capacity restoration decoder can more effectively reconstruct information for missing modalities, thereby enhancing the overall system performance.

%%%%%%%%%%%%%%%%%%%%%%%%%%%%%%%%%%%%%%%%%%%%%%%%%%%%%%%%%%%%%%%%%%%%%%%%%%%%%%%%%%%%%%%%%%%%%%%%%%%%%%%%%%%%%%%%%%%%%%%%%%%%%%%%%%%%%%%%%
% Table 6
\begin{table}[htbp]
\centering
\scriptsize
\def\arraystretch{0.75}
\begin{tabular}{cc  cc  cc  cc}
\toprule
\multirow{2}{*}{\shortstack {\\ {\textit{Dim}}}} & \multirow{2}{*}{\shortstack {\\ {\textit{Depth}}}} 
& \multicolumn{2}{c}{\textit{Modality 1}} 
& \multicolumn{2}{c}{\textit{Modality 2}} 
& \multicolumn{2}{c}{\textit{Full}} \\
\cmidrule(lr){3-4} \cmidrule(lr){5-6} \cmidrule(lr){7-8}
& & ACC & AUC & ACC & AUC & ACC & AUC \\
\midrule
\multirow{3}{*}{128}
& 4 & 75.29 & 92.40 & 76.86 & 93.54 & 77.37 & 93.88 \\
& 6 & 75.19 & 92.67 & 76.31 & 93.76 & 77.99 & 94.21 \\
& 8 & 75.48 & 92.29 & 76.80 & 93.89 & 77.95 & 94.27 \\
\midrule
\multirow{3}{*}{256}
& 4 & 75.36 & 92.38 & 76.62 & 93.67 & 77.87 & 94.03 \\
& 6 & 75.42 & 92.58 & 77.33 & 93.81 & 77.98 & 94.13 \\
& 8 & 75.48 & 92.02 & 77.48 & 93.68 & 77.82 & 94.05 \\
\midrule
\multirow{3}{*}{\textbf{512}}
& 4 & 75.59 & 92.43 & 77.16 & 93.62 & 78.04 & 94.19 \\
& 6 & 75.69 & 92.49 & 77.12 & \textbf{93.88} & 78.21 & 94.11 \\
& \textbf{8} & \textbf{76.51} & \textbf{92.70} & \textbf{77.77} & 93.85 & \textbf{78.83} & \textbf{94.46} \\
\bottomrule
\end{tabular}

\caption{Performance of the restoration decoder based on dimensions and depth.}
\label{table:table_6}
\end{table}
%%%%%%%%%%%%%%%%%%%%%%%%%%%%%%%%%%%%%%%%%%%%%%%%%%%%%%%%%%%%%%%%%%%%%%%%%%%%%%%%%%%%%%%%%%%%%%%%%%%%%%%%%%%%%%%%%%%%%%%%%%%%%%%%%%%%%%%%%

\section{Discussion and Conclusion}
This study proposed PhysioME, a physiological signal-specific framework designed to address the challenges of missing modalities frequently encountered in real-world clinical environments. Building upon SynthSleepNet \cite{lee2025toward}, an SSL framework based on MultiMAE \cite{bachmann2022multimae}, PhysioME is specifically designed to capture the temporal characteristics of physiological signals and heterogeneity across sensor types, enabling robust handling of various missing modality scenarios.

Leveraging this architectural design, PhysioME demonstrated strong predictive performance and generalization across diverse missing modality settings owing to several key structural components. First, the DP-NeuroNet serves as a core component in the success of PhysioME. By applying two different temporal augmentations to a dual-path network that shares weights across identical NeuroNet instances, DP-NeuroNet effectively captures the temporal dynamics of physiological signals and significantly enhances the quality of the modality-specific representations. These high-dimensional representations play a crucial role in improving both the prediction accuracy and generalization capability (Table \ref{table:table_3}). Second, the restoration decoder enables semantically informed reconstruction by modeling cross-modal interactions rather than relying on simple imputation methods such as mean substitution or input replication. In particular, the restoration decoder is optimized using the stop-gradient technique, which receives input from the multimodal encoder while preventing the gradients from flowing backward. This allows the decoder to focus solely on the restoration tasks. In addition, the decoder benefits from a deeper architecture that facilitates effective learning (Tables \ref{table:table_5} and \ref{table:table_6}).

In clinical practice, it is often difficult to acquire all the physiological signals owing to various constraints. For example, EEG acquisition may be physically restricted during brain surgery \cite{bitar2024utility} and ABP sensors may not be applicable to patients receiving anticoagulants or those with impaired clotting function \cite{puckett2003monitoring}. Similarly, severe motion artifacts can compromise the reliability of ECG signals, and PPG measurements may be unreliable under conditions such as hypothermia or peripheral vasoconstriction \cite{littmann2021electrocardiographic}. Despite these limitations, PhysioME consistently maintains predictive stability and robustness, demonstrating its potential to reliably support clinical decision-making in real-world healthcare environments.

However, this study has several limitations. First, the performance was evaluated under a pre-defined set of modality combinations. Generalization to entirely new combinations or domain transfer scenarios requires further investigation. Second, the architectures of DP-NeuroNet and the restoration decoder are relatively complex in terms of the training efficiency. Specifically, a restoration decoder must be instantiated for each new modality to introduce scalability challenges. Future work will evaluate cross-hospital generalization using multi-center datasets collected from diverse institutions and patient populations, and further assess robustness not only to missing modalities but also to noise and artifacts commonly observed in clinical data.

\section{Appendix}
\subsection{Background: NeuroNet}
NeuroNet \cite{lee2024neuronet} is a self-supervised learning (SSL) framework that combines contrastive learning and masked prediction tasks to extract generalizable representations from unlabeled input data. In this study, NeuroNet was adopted as the core component to extract modality-specific features from each physiological signal.

The NeuroNet encoder is built around a frame network module. To effectively process time-series data, this module comprises a shared convolutional block and three parallel feature extractors with kernel sizes of 3, 5, and 7, each of which follows a convolutional residual-block architecture. The feature vectors are concatenated and passed through an MLP layer, producing frame-wise vectors.

While performing the masked prediction task based on the MAE strategy, a subset of the frame-wise vectors is randomly selected and input into a ViT-based encoder along with a class token. The encoder generates a latent representation from the observed frames, and the decoder reconstructs the masked frames using the latent representation and mask tokens. The reconstruction loss is calculated only over the masked frames using the MSE loss:

\begin{equation}
\mathcal{L}_{\text{inter\_recon\_loss}} = \frac{1}{|\mathcal{M}|} \sum_{m \in \mathcal{M}} \left\| r^{(m)} - z^{(m)} \right\|_2^2
\end{equation} where \(\mathcal{M}\) denotes the set of masked frame indices. The vectors $r$ and $m$ correspond to the outputs of the decoder and the frame network, respectively.

In addition, NeuroNet employs contrastive learning based on the NT-Xent loss to ensure consistency across different views of the same input sequence. Two random samples are generated from a single input sample, and each sample is processed using the encoder and an MLP to produce the latent vectors \(\bm{z}_i\) and \(\bm{z}_j\). The NT-Xent loss is defined as:

\begin{equation}
\mathcal{L}_{\text{NT-Xent\_loss}} = -\log \frac{\exp\left(\text{sim}(\bm{z}_i, \bm{z}_j)/\tau\right)}{\sum_{k=1}^{2N} \mathbf{1}_{[k \neq i]} \exp\left(\text{sim}(\bm{z}_i, \bm{z}_k)/\tau\right)},
\end{equation} where \(\text{sim}(\cdot)\) denotes the cosine similarity and \(\tau\) is the temperature parameter.

NeuroNet performs two independent masked prediction tasks using the same encoder and decoder. As a result, the total loss function combines two reconstruction losses and one contrastive loss:

\begin{equation}
\mathcal{L}_{\text{total}} = \frac{1}{2} \left( \mathcal{L}_{\text{inter\_recon\_loss}_1} + \mathcal{L}_{\text{inter\_recon\_loss}_2} \right) + \alpha \mathcal{L}_{\text{NT-Xent\_loss}},
\end{equation} where \(\alpha\) is a hyperparameter that balances the reconstruction and contrastive losses.

\subsection{Hyperparameter Setting}
The experimental environment was a machine equipped with an Intel Xeon Gold 6338 CPU 2.00GHz, 256GB of RAM, and two NVIDIA GPU Ada 6000. The codes, including data preprocessing and model development, were implemented based on Pytorch 2.7.1 and Python 3.11. The hyperparameter settings are listed in Appendix Table \ref{table:hyperparameters}.

\begin{table}[http]
\renewcommand{\arraystretch}{0.9}
\centering
\scriptsize
\begin{tabular}{ccc}
\hline
\multicolumn{1}{c}{}                      & \multicolumn{1}{c}{\textit{\begin{tabular}[c]{@{}c@{}}\textbf{Sleep Stage}\\ \textbf{Classification}\end{tabular}}} & \textit{\begin{tabular}[c]{@{}c@{}}\textbf{Hypotension}\\ \textbf{Prediction}\end{tabular}} \\ \hline
\multicolumn{3}{c}{\textit{DP-NeuroNet}}                                                                                                                                                                                    \\ \hline
epoch                 & 50 & 50 \\
sampling rate         & 100 & 100 \\
batch size            & 128 & 128 \\
frame size (sec)      & 4 & 3 \\
overlap step (sec)    & 1 & 3 \\
encoder dim           & 512 & 512 \\
encoder depth         & 8 & 8 \\
encoder head          & 6 & 6 \\
decoder dim           & 256 & 256 \\
decoder depth         & 8 & 8 \\
decoder head          & 4 & 4 \\
mask ratio            & 0.8 & 0.8 \\
balance scale         & 1.0 & 1.0 \\
projection hidden     & {[1024, 512]} & {[1024, 512]} \\
optimizer             & AdamW & AdamW \\
optimizer momentum    & (0.9, 0.999) & (0.9, 0.999) \\
learning rate         & 1e-05 & 1e-05 \\ \hline
\multicolumn{3}{c}{\textit{PhysioME}} \\ \hline
physiological signals & Fpz-Cz / Pz-Cz / EOG & ABP / ECG / PPG \\
epoch                 & 50 & 50 \\
batch size            & 512 & 512 \\
encoder dim           & 512 & 512 \\
encoder depth         & 6 & 4 \\
encoder head          & 8 & 8 \\
decoder dim           & 256 & 256 \\
decoder depth         & 4 & 3 \\
decoder head          & 8 & 8 \\
recon decoder dim     & 256 & 256 \\
recon decoder depth   & 8 & 8 \\
recon decoder head    & 8 & 8 \\
projection hidden     & {[1024, 512]} & {[1024, 512]} \\
temperature scale     & 0.1 & 0.1 \\
mask ratio            & 0.4 & 0.4 \\
optimizer             & AdamW & AdamW \\
optimizer momentum    & (0.9, 0.999) & (0.9, 0.999) \\
learning rate         & 2e-4 & 2e-4 \\
balance scale         & $\alpha$: 1.0 / $\beta$: 1.0 / $\gamma$: 1.0 & $\alpha$: 1.0 / $\beta$: 1.0 / $\gamma$: 1.0 \\
lora (r)                & 4 & 4 \\
lora (alpha)            & 16 & 16 \\
lora (dropout)          & 0.05 & 0.05 \\ \hline
\multicolumn{3}{c}{\textit{Downstream Task}} \\ \hline
epoch                 & 20 & 20 \\
batch size            & 512 & 512 \\
optimizer             & AdamW & AdamW \\
optimizer momentum    & (0.9, 0.999) & (0.9, 0.999) \\
learning rate         & 1e-05 & 1e-05 \\ \hline
\end{tabular}
\caption{Hyperparameter Settings}
\label{table:hyperparameters}
\end{table}

\newpage
\bibliography{aaai2026}
\end{document}